%% file: icsoc23.tex
\documentclass[runningheads]{llncs}
\usepackage{graphicx}

\usepackage{cite}
\usepackage{algorithmic}
\usepackage{graphicx}
\usepackage{xcolor}
\usepackage{url}
\usepackage{hyperref}
  
\usepackage[inline]{enumitem}

\newcommand{\as}{adaptive system}

\newcommand{\app}{\begin{small}\textsf{Chat4XAI}\end{small}}
\newcommand{\runin}[1]{\noindent\textbf{#1.}}
\newcommand{\runintext}[1]{\noindent\textbf{#1}}

\begin{document}

\title{An AI Chatbot for Explaining Deep Reinforcement Learning Decisions of Service-oriented Systems}

\author{Andreas Metzger \orcidID{0000-0002-4808-8297} \and Jone Bartel \and Jan Laufer}
\institute{paluno -- The Ruhr Institute for Software Technology\\
University of Duisburg-Essen, Essen, Germany\\
\email{firstname.lastname@paluno.uni-due.de}
}

\maketitle

\begin{abstract}
Deep Reinforcement Learning (Deep RL) is increasingly used to cope with the open-world assumption in service-oriented systems.
Deep RL was successfully applied to problems such as dynamic service composition, job scheduling, and offloading, as well as service adaptation.
While Deep RL offers many benefits, understanding the decision-making of Deep RL is challenging because its learned decision-making policy essentially appears as a black box.
Yet, understanding the decision-making of Deep RL is key to help service developers perform debugging, support service providers to comply with relevant legal frameworks, and facilitate service users to build trust.
We introduce Chat4XAI to facilitate the understanding of the decision-making of Deep RL by providing natural-language explanations.
Compared with visual explanations, the reported benefits of natural-language explanations include better understandability for non-technical users, increased user acceptance and trust, as well as more efficient explanations.
Chat4XAI leverages modern AI chatbot technology and dedicated prompt engineering.
Compared to earlier work on natural-language explanations using classical software-based dialogue systems, using an AI chatbot eliminates the need for eliciting and defining potential questions and answers up-front.
We prototypically realize Chat4XAI using OpenAI's ChatGPT API and evaluate the fidelity and stability of its explanations using an adaptive service exemplar.

\keywords{chatbot \and explainable AI \and reinforcement learning \and service engineering \and service adaptation}
\end{abstract}

\input{intro}

\input{poc}

\input{evaluation}

\input{discussion}

\input{sota}
\input{summary}

\bibliographystyle{splncs04}
\bibliography{main,chat4xai}

\end{document}

%% file: intro.tex
\section{Introduction}
\label{sec:intro}

Reinforcement Learning (RL) is increasingly used to cope with the open-world assumption of service-oriented systems, as it helps to address the design-time uncertainty during service and systems engineering~\cite{MetzgerEtAl2022,BaresiNG06}.
In general, RL aims to learn an optimal decision-making policy for executing a suitable action in a given environment~\cite{sutton_reinforcement_2018}.
In a service-oriented system, RL can learn suitable actions via the service-oriented system's interactions with its initially unknown environment and thereby can make use of information only available at runtime~\cite{CAISE2020}. 
RL helped successfully address various problems in service-oriented systems, including 
dynamic service composition~\cite{RazianFBTB22},
task/job scheduling~\cite{JamilISMB22,HuangWMCC22,YuZSLLWWL22}, 
resource management~\cite{IftikharEa23},
and service adaptation~\cite{MutanuK19}.

\runin{Need for Explainability} Recent research on using RL for realizing service-oriented systems leverages Deep RL algorithms, which represent their decision-making policy as a deep neural network (examples include~\cite{MaXu2023,HuangWMCC22,MoXZQL21,GhanadbashiEa2023}).
Deep RL inputs are not limited to elements of finite or discrete sets, and the used neural networks can generalize well over unseen inputs.
Deep RL can also natively capture concept drift in a service-oriented system's environment without the need to explicitly introduce mechanisms to monitor such concept drift~\cite{CAISE2020}.

However, one key shortcoming of Deep RL is that the learned decision-making policy is not represented explicitly, but is hidden in the parametrization of the neural network. 
For service developers and users, the decision-making of Deep RL thus essentially appears as a black box~\cite{puiutta_explainable_2020}.
This means that we require techniques to explain and interpret the internal workings of such black-box systems and how their decisions are made~\cite{MohseniZR21,CamilliMS22,miller_explanation_2019,GuidottiMRTGP19}.

Explaining the decision-making of Deep RL can help service developers debug the reward function by understanding why Deep RL took certain decisions.
The successful application of Deep RL depends on how well the learning problem, and in particular the reward function, is defined~\cite{Dewey14}. 
Further, explainability can facilitate regulatory compliance~\cite{MohseniZR21}.
For example, in the EU, service providers must ensure that their services comply with the relevant legal frameworks, such as the General Data Protection Regulation and the forthcoming AI Act.
Third, explanations facilitate  service users to build trust.
They can understand how the service arrived at its results and thus can accept its results or not~\cite{MohseniZR21}.

\runin{Problem Statement} 
Two major types of explanation formats can be distinguished~\cite{MohseniZR21,MalandriMMN23,CambriaMMMN23}: (\emph{i}) visual explanations, including graphical user interfaces, charts, data visualization, or heatmaps, and (\emph{ii}) verbal explanations, which, for instance, may take the form of a natural-language dialogue between the explainer and explainee.
The chosen presentation method has a direct effect on user comprehension and, therefore, on the success of the explanation~\cite{MalandriMMN23}.
Compared with visual explanations, the benefits of verbal explanations reported in the literature~\cite{MariottiAG20,CambriaMMMN23} include (1) better understandability for people with diverse backgrounds as well as non-technical users, (2) increased user acceptance and trust, and (3) more efficient explanations.

While the literature on using Deep RL for service-oriented systems provides extensive and systematic evaluations of the performance of Deep RL~\cite{MaXu2023,HuangWMCC22,MoXZQL21,GhanadbashiEa2023}, the problem of how to explain the decision-making of Deep RL using natural language was not yet addressed.
In the broader area of explainable AI (XAI), approaches for providing natural-language explanations for machine learning exist~\cite{KuzbaB20,LiaoGM20,Nguyen2022,MalandriMMN23}.
However, these XAI approaches focus on supervised learning and not on RL.
Also, they are all built using classical software-based dialogue systems~\cite{MotgerFM23}, which require the additional engineering step of eliciting and defining potential questions and answers up-front.

\runin{Contributions}
We introduce \app{}, which leverages the capabilities of a modern AI chatbot powered by a large language model to provide natural-language explanations about the decision-making of Deep RL.
AI chatbots are intriguing in that they provide natural-language answers to any natural-language question posed to them.
However, as a downside of this flexibility, the underlying large language model may "hallucinate", i.e., generate nonsensical text unfaithful to the provided source input~\cite{JiLFYSXIBMF23}.
This means AI chatbots may deliver explanations that do not faithfully explain the decision-making of RL, i.e., the explanations may exhibit low fidelity~\cite{GuidottiMRTGP19}.
In addition, AI chatbots may provide different explanations for the very same question asked, i.e., the explanations may exhibit low stability~\cite{Robnik-SikonjaB18}.

To deliver natural-language explanations with high fidelity and high stability, \app{} uses dedicated prompt engineering for the AI chatbot and careful selection of the hyper-parameters of the underlying large language model.
Prompt engineering helps increase the correctness of the answers by providing a set of targeted, initial questions (a.k.a. prompts) before the actual question~\cite{WhiteEA2023,StrobeltWSHBPR23}.

We prototypically realize and evaluate \app{} using OpenAI's ChatGPT Completion API.
We evaluate the fidelity and stability of the explanations delivered by \app{} using an adaptive cloud service exemplar realized using Double DQN as Deep RL algorithm.
We assess \app{} for different prompting strategies, open as well as closed questions, and different hyper-parameter settings.
To contextualize the performance of \app{}, we compare \app{}'s explanations with how well human software engineers were able to understand the decision-making of Deep RL using visual explanations~\cite{TAAS2023}.

\runin{Paper Structure} Sect.~\ref{sec:chat4xai} describes the conceptual architecture and proof-of-concept implementation of \app{}.
Sect.~\ref{sec:evaluation} provides the experiment design and results.
Sect.~\ref{sec:future} discusses limitations and future enhancements.
Sect.~\ref{sec:sota} relates \app{} to existing work.

%% file: poc.tex
\section{Chat4XAI Architecture and Realization}
\label{sec:chat4xai}

\begin{figure}[t]
  \centering
   \includegraphics[width=.85\columnwidth]{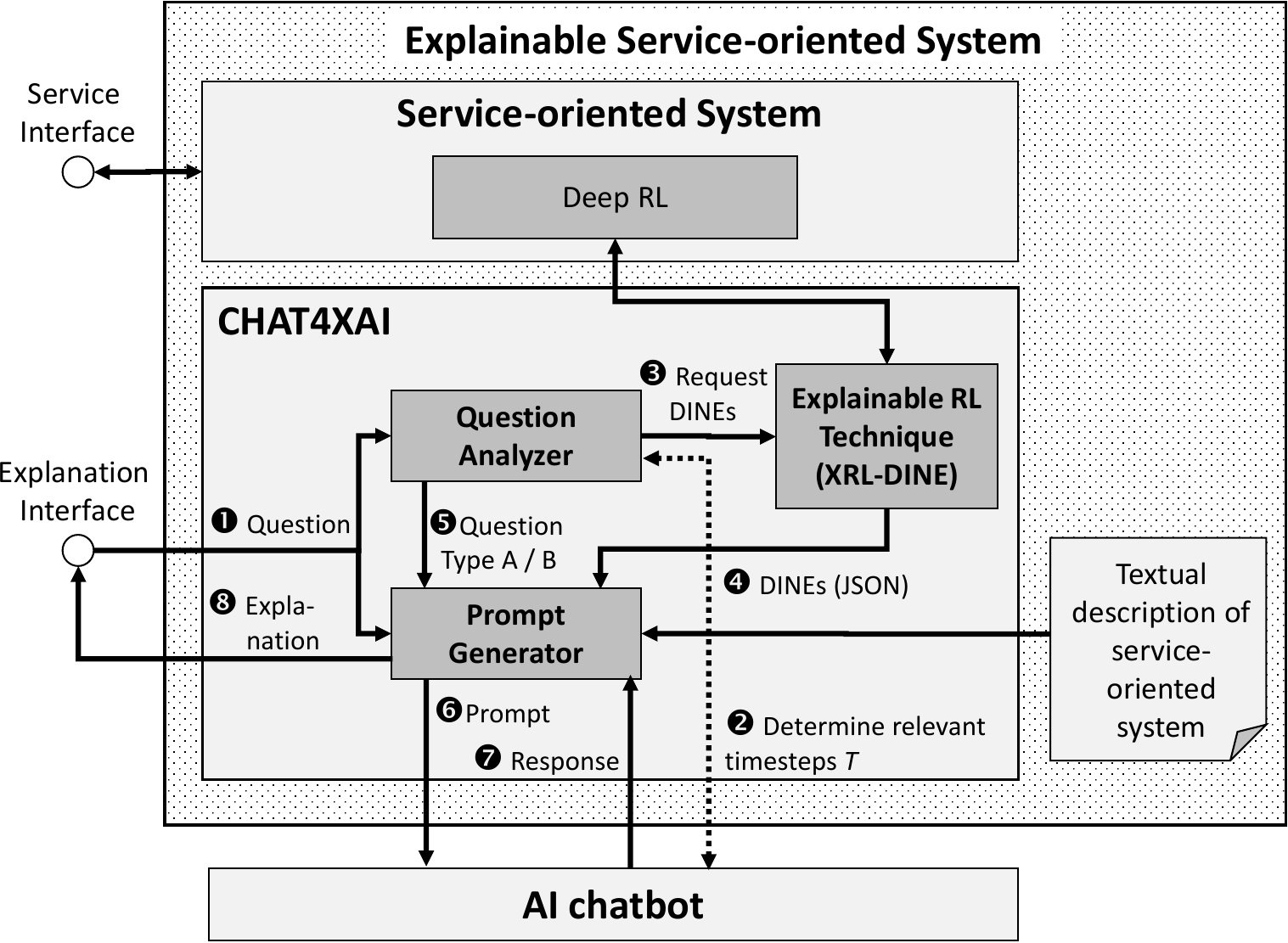}
  \caption{\app{} conceptual architecture and embedding into service-oriented system}
  \label{fig:chat4xai}
  \vspace{-1.5em}

\end{figure}

Fig.~\ref{fig:chat4xai} shows the main conceptual components of \app{} and how they may be embedded to realize an explainable service-oriented system.
Fig.~\ref{fig:chat4xai} also shows the control and data flow among these components and in which order this happens (shown as numbers in black circles).
\app{} leverages the output of XRL-DINE, a state-of-the-art explanation technique~\cite{ACSOS22}, to generate natural-language explanations.
Below we explain the different conceptual components of \app{} followed by how we prototypically realized \app{} via ChatGPT.

\subsection{XRL-DINE}
\label{sec:dines}
XRL-DINE generates different types of so-called \emph{Decomposed Interestingness Elements} (DINEs), which provide insights into the decision-making of Deep RL.
In the original approach in~\cite{ACSOS22}, DINEs are visualized in the XRL-DINE graphical user interface.
Here, we use the information of the DINEs as input to \app{}.

XRL-DINE combines and enhances two explanation techniques: 
\emph{Interestingness Elements} facilitate identifying situations where the decision-making of Deep RL is uncertain.
It thereby helps select interesting decisions among the many decisions taken by Deep RL at runtime~\cite{sequeira_interestingness_2020}.
Situations, where Deep RL is uncertain, may point to states which require more training, or where the reward function may need to be re-engineered to provide stronger rewards.
\emph{Reward Decomposition} splits the reward function into sub-functions, called reward channels, each of which reflects a different aspect of the learning goal~\cite{juozapaitis_explainable_2019}.
For each reward channel, a separate decision-making policy is learned.
To select a concrete action, the individual decision-making policies are combined and an action is selected from this combined decision-making policy. 
The trade-offs between the different learning goal become observable via these reward channels. 

Technically, the information of the DINEs is encoded in JSON format.
JSON is an open-standard data interchange format using human-readable text, and thus can serve directly as input to the AI chatbot. 
To illustrate, the JSON snippet below gives an example of the information contained within a ``Relative Reward Channel Dominance'' DINE.
This type of DINE provides the actual contribution of each reward channel to the aggregated decision.
\vspace{-.5em}
\noindent{\footnotesize\begin{verbatim}
    "Action 1": { "Reward Channel A": 0.35, 
                  "Reward Channel B": 2.61, 
                  "Reward Channel C": 1.19 },
    "Action 2": { "Reward Channel A": 0.13, 
                  "Reward Channel B": 0.0, 
                  "Reward Channel C": 1.01 }, ...
\end{verbatim}}
\vspace{-.5em}
In this example, \emph{Action 1 }is chosen, as it has the highest relative reward with \emph{Reward Channel B }contributing most to the aggregate decision.

\subsection{Question Analyzer}
The Question Analyzer classifies a given question into one of two question types:

\runintext{Question Type A} concerns a \emph{single} decision, i.e., it covers the decision taken at a single timestep. 
One example is to ask why Deep RL decided for adaptation $X$ rather than $Y$ at timestep $t$.

\runintext{Question Type B} concerns a sequence of decisions, i.e., it covers the decision trajectory of several timesteps.
One example is to ask how often, along the concerned timesteps $t_k$ to $t_l$, Deep RL was uncertain in its decisions.

The reason for classifying a question into one of these two types is twofold.
First, while state-of-the-art AI chatbots are capable of directly answering natural-language questions (so-called 'zero-shot' prompting), the quality of the answers can be increased via prompt engineering, i.e., by providing a set of initial questions (a.k.a. prompts) before the actual question to be answered~\cite{WhiteEA2023,StrobeltWSHBPR23}.
Differentiating between the two types of questions allows for a more targeted prompt engineering. 
Second, AI chatbots typically limit the cumulative length of questions and answers per conversation (see Section~\ref{sec:poc}).
Depending on the question type, we can filter the number and types of relevant DINEs provided as input to the AI chatbot to remain within length limits.

To identify the question type, we ask the AI chatbot to provide us with a list $T$ of all relevant timesteps mentioned in the question.
Then, by counting the size of $T$ we can identify whether it is a question of Type A ($|T| = 1$) or Type B ($|T| > 1$).
Note that in the case of $|T| = 0$, one may use a default set of timesteps; e.g., the 20 most recent ones, i.e., $T = (t_{now-20}, \ldots, t_{now})$.

Depending on the question type, the DINEs for a single time step (Type A questions) or for multiple time steps (Type B questions) are requested from XRL-DINE and forwarded to the Prompt Generator.

\subsection{Prompt Generator}
The Prompt Generator receives the following pieces of information:
\begin{itemize}
\item A textual \emph{description of the service-oriented system} providing relevant concepts, including domain-specific terms, the different actions available, and the learning goals.
These concepts provide relevant service-specific knowledge to the AI chatbot, as the DINEs also refer to these concepts.
If \app{} is used to deliver an explanation interface to service users (as depicted in Figure~\ref{fig:chat4xai}), this description is provided by the service developer or provider.
\item The \emph{question type} as identified by the Question Analyzer.
\item The \emph{DINEs} in JSON format for the relevant timesteps $T$.
\item The \emph{actual question}, which is provided by the explainee.
\end{itemize}

The Prompt Generator generates and issues the following prompts:

\runintext{Prompt 1} provides relevant concepts to the AI chatbot by providing it with the textual description of the service-oriented system, introduced by the text\\ {\footnotesize\tt
The following scenario description will be available...}

\runintext{Prompt 2} provides the context for the answers to be given by the AI chatbot.
It gives the type of data that will follow after this prompt (which depends on the question type) together with a reference to the service-oriented system to allow the AI chatbot to connect to the description given in Prompt 1:
\begin{itemize}
\item \textit{Question Type A:} {\footnotesize\tt
You will be given the state for a single timestep of <name from scenario> as JSON enclosed in ***:}
\item \textit{Question Type B:}
{\footnotesize\tt
You will be given a trajectory of timesteps for <name> \\as JSON enclosed in ***:}
\end{itemize}

\runintext{Prompt 3} provides the actual DINEs in JSON format enclosed in {\footnotesize\tt***} to mark the boundaries between the DINEs and the final question in Prompt 4.

\runintext{Prompt 4} provides the actual question of the explainee.

Typically, an explainee will ask open questions (i.e., without any fixed set of answers to choose from) and expect an answer in concrete terms.
Directly asking open Type B questions turned out to be a too challenging for the AI chatbot.
We thus employ "chain of thought" prompt engineering, in which the AI chatbot is given smaller subtasks leading up to the final results.
In a first prompt, we ask the AI chatbot to provide us with a list of relevant timesteps; e.g., the ones where Deep RL is uncertain.
In a second prompt, we then ask the AI chatbot concrete questions about these relevant timesteps; e.g., to count these.

\subsection{Proof-of-concept Realization}
\label{sec:poc}
We developed our proof-of-concept realization of \app{} in Python using the OpenAI ChatGPT Completion API\footnote{\url{https://platform.openai.com/docs/api-reference}}.
We are using GPT 3.5 turbo as a large language model, which offers a faster generation of responses, and is capable of providing more in-depth answers.
The following hyper-parameters are considered in our realization and experiments:
\vspace{-.8em}
\begin{itemize}
\item \emph{n} gives how many responses to generate for each prompt.
\item \emph{max\_token} limits the length of the answer.
The OpenAI API imposes two constraints on the overall length of a request, i.e., a sequence of prompts and responses.
First, there is an overall token limit of $4,096$ tokens (ca. 3,000 words) per request. 
Second, there is a limit of $90,000$ tokens per minute.
\emph{max\_token} helps to set a trade-off between these two constraints.
\item \emph{temperature} controls text generation behavior.
Temperature $\in [0, 2]$ controls the randomness of the text, with a higher temperature resulting in more "creative" text (but with a higher risk of hallucinations).
A temperature of 0 leads to deterministic text generation behavior.
\item \emph{Top\_p sampling} is an alternative to temperature.  
Instead of considering all possible tokens, only a subset of tokens is considered whose cumulative probability adds up to top\_p $\in [0, 1]$.
\end{itemize}

Table~\ref{tab:examples} gives examples of explanations generated via the proof-of-concept realization of \app{}.

\begin{table*}[h!]
	\caption{Examples for natural-language explanations generated by \app{}}
\centering
\includegraphics[width=1\columnwidth]{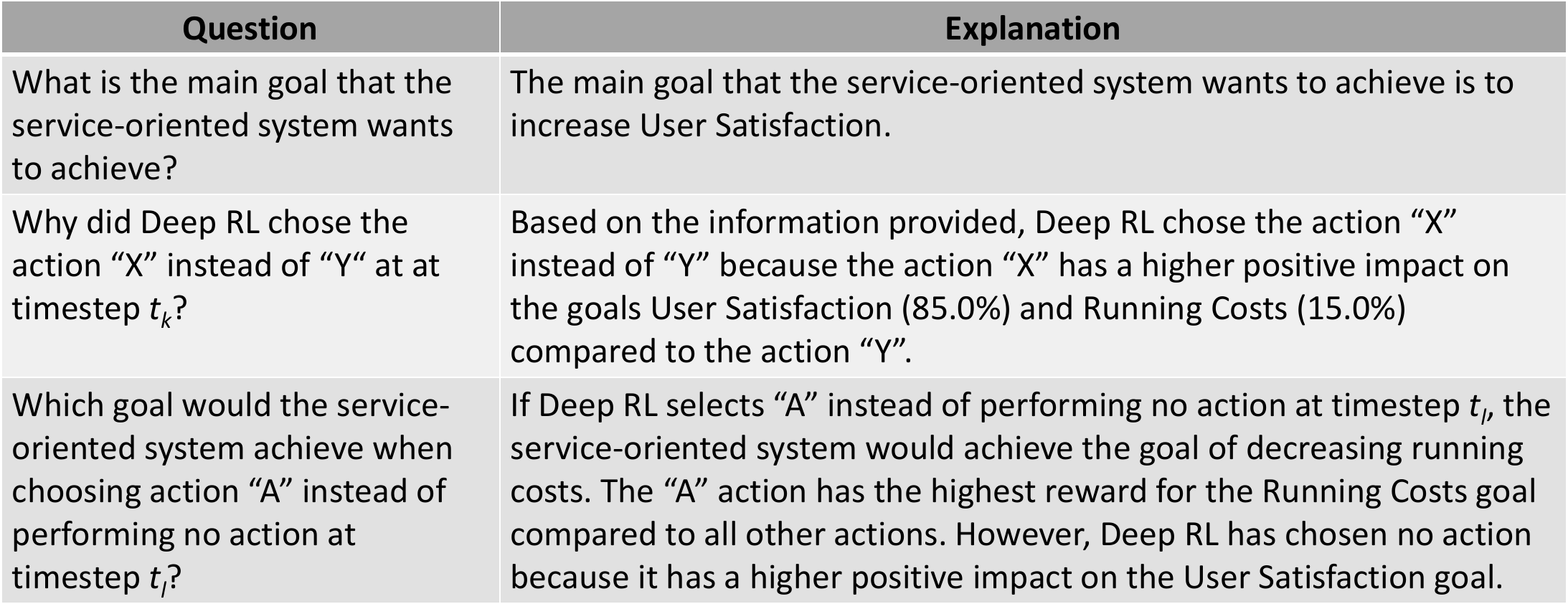}

	\label{tab:examples}

\end{table*}

%% file: evaluation.tex
\section{Experiments}
\label{sec:evaluation}

We perform a set of controlled experiments to evaluate the fidelity and stability of \app{}'s explanations depending on different configurations of our explanation technique.
We also compare the results of \app{} with the results of an empirical user study from the literature~\cite{TAAS2023}.
This user study assessed how well software engineers were able to understand the decision-making of Deep RL by only using explanations in visual form, i.e., using the XRL-DINE graphical user interface.
To facilitate reproducibility, relevant background and supplementary material is available from \url{https://gitlab.com/xrl2/Chat4XAI/}.

\subsection{Metrics}
We chose the following metrics from the explainable AI literature for assessing \app{}~\cite{Robnik-SikonjaB18,GuidottiMRTGP19}:

\runintext{Fidelity} expresses how well the explanations reflect the behavior of the black-box model.
As we use AI chatbots to generate explanations, \app{} exhibits the risk of hallucination, i.e., producing answers not corresponding to the learned Deep RL decision-making policy.
We quantify fidelity by measuring the rate of correct explanations for $m$ given questions.
Let $x_i = 1$ mean correct explanation, and 0 otherwise, then fidelity is computed as $ \sum x_i / m $.

\runintext{Stability} gives the degree to which the same explanation is generated for the same input. 
Explanations of \app{} for the same input may vary because rge AI chatbot may produce non-deterministic results depending on the chosen hyper-parameter settings.
We measure stability as $1-\sigma$, with $\sigma$ being the standard deviation of fidelity across several experiment repetitions for the same input and configuration of independent variables (see Section~\ref{sec:setup}).

\vspace{1em}

We seek to directly compare the performance of \app{} with the performance of software engineers using XRL-DINE.
We, therefore, selected the following metric from the user study of XRL-DINE~\cite{TAAS2023}:

\runintext{Effectiveness} quantifies the performance of software engineers by measuring the rate of correctly answered questions concerning Deep RL decision-making among $n$ participants.
Let $c_i = 1$ mean correct explanation, and 0 otherwise, then the effectiveness is computed as $ \sum c_i / n $.

\subsection{Experiment Environment}
\label{sec:exemplar}
To be able to compare the results for \app{} with the results of the user study reported in~\cite{TAAS2023}, we use the same environment as in the user study.
In particular, this means we use the same set of DINEs covering the same 21 timesteps concerning the decision-making of the chosen service-oriented system exemplar realized using Double DQN with Experience Replay as Deep RL algorithm.

The service-oriented system exemplar SWIM represents an adaptive multi-tier webshop~\cite{moreno_swim_2018}. 
SWIM simulates an actual web shop, while allowing to speed up the simulation to cover longer periods of time.
The goal of adaptation in SWIM is to maximize a given utility function in the presence of varying workloads.
SWIM can be adapted by (1) adding / removing web servers, and (2) changing the proportion of requests for which optional, computationally intensive recommendations are generated. 
While both types of adaptations have an impact on user satisfaction (due to their influence on throughput and response time), adaptations of type (1) have an impact on costs (due to the costs of more/fewer servers), and adaptations of type (2) have an  impact on revenue (due to recommendations leading to potential further purchases in the webshop).
We use the same decomposed reward function as introduced in~\cite{ACSOS22}, which aims to balance the aforementioned conflicting QoS goals via the weights $a, b,$ and $c$.

\begin{center}
\vspace{-.5em}
$
R_{\mathrm{total }}=
a \cdot R_{\mathrm{user\_satisfaction}} + 
b \cdot R_{\mathrm{revenue }} + 
c \cdot R_{\mathrm{costs }}
$
\end{center}

\runin{User Study} 
Below, we provide key aspects of the user study from~\cite{TAAS2023} to provide a context for the results reported in Section~\ref{sec:results}.
The user study involved 54 software engineers from academia (82\%) and industry (8\%), most of which held an academic degree in software engineering or related fields (35\% Bachelor's degree, 41\% Master's degree, 11\% PhD).  
The user study was carried out using an online questionnaire. 
Study participants were asked eight closed questions (see Section~\ref{sec:setup}) and were provided with several single-choice answers for each question.
Study participants also received a description of the service-oriented system, which included: (1)
an explanation of the SWIM exemplar; (2) a list of the QoS goals, among which Deep RL should seek a trade-off; (3) a list of possible adaptations (i.e., RL actions), together with examples of typical effects of these adaptations.

\subsection{Experiment Setup}
\label{sec:setup}

\runin{Independent Variables} To analyze the performance of \app{}, we varied the following three main independent variables.

\emph{Prompting:} We analyze the performance depending on whether (\emph{i}) zero-shot prompting or (\emph{ii}) prompt engineering is used.
This indicates how much \app{}'s performance is impacted by prompt engineering, resp. how robust the approach would be independent of any specific prompt engineering.

\emph{Question Form:} We analyze the performance of \app{} in providing explanations for open questions (which would be the typical scenario in a practical setting) and closed questions (like in the user study from~\cite{TAAS2023}).

\emph{Hyper-parameters:} 
We analyze the performance of \app{} for different concrete settings of temperature and top\_p.
We vary temperature $\in \{0, 0.2, 0.5, 1\}$.
For each temperature, we vary top\_p $\in \{1, 0.8, 0.5\}$ and report the aggregated results per 
temperature.
We use 54 repetitions to get the same number of explanations as in the user study.
We split these 54 repetitions into three clusters of 18 repetitions, with each cluster having a different top\_p setting.
To efficiently execute the experiment, we set $n = 18$, i.e., retrieve 18 answers per prompt.
Also, we set \emph{max\_token} $=350$, delivering a good trade-off between length of answers and throughput.

\runin{Questions to be Explained}
To assess the performance of \app{}, we have to choose concrete questions that are posed to the AI chatbot to retrieve explanations.
As we are interested in comparing the performance of \app{} with that of software engineers from the empirical user study in~\cite{TAAS2023}, we use the same set of questions that were formulated there.
These questions are shown in Table~\ref{tab:questions}.

\begin{table*}[t!]
    \centering
   	\caption{Questions for which explanations should be given\protect\footnotemark}
    \label{tab:questions}
	\includegraphics[width=1\textwidth]{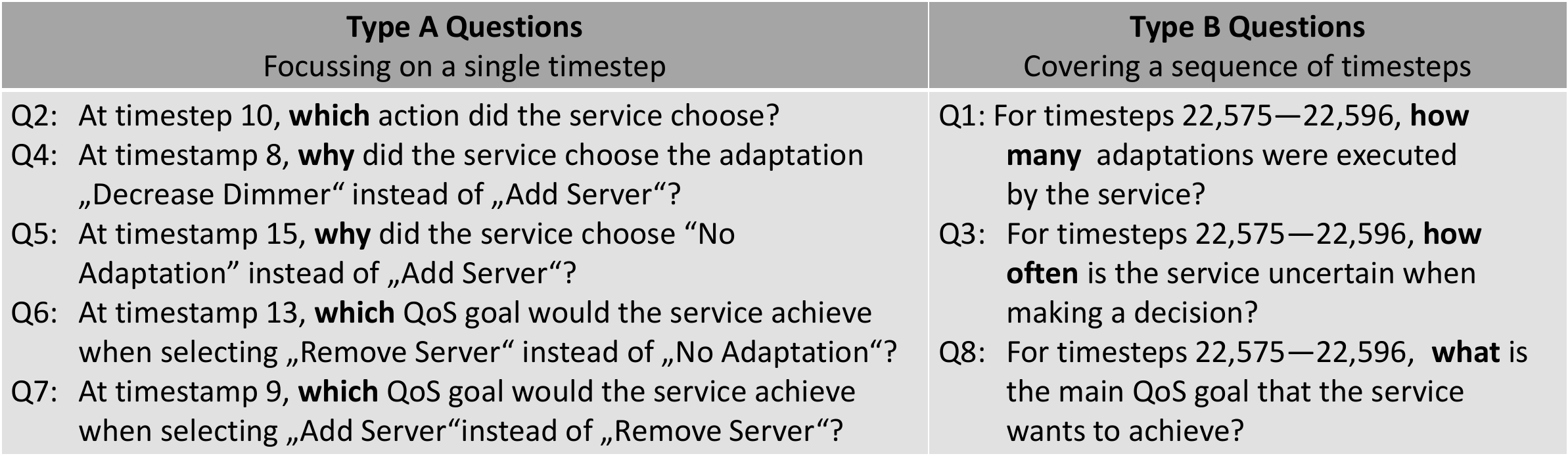}         
\end{table*}

\footnotetext{Questions are numbered in the order they were asked user study participants. Question text adapted from~\cite{TAAS2023} and edited for clarity.}


\subsection{Experiment Results} 
\label{sec:results}
Table~\ref{tab:results} presents the overall results concerning the fidelity and stability of \app{} as well as the effectiveness of software engineers from the user study.

\begin{table*}[t!]
	\caption{Experiment Results}
\centering
\includegraphics[width=1\columnwidth]{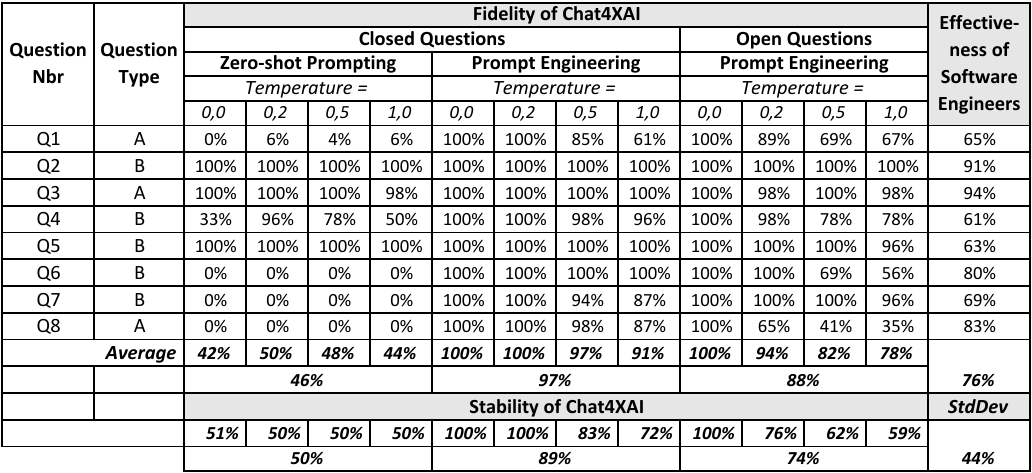}

	\label{tab:results}

\end{table*}

As expected \app{} with prompt engineering outperforms \app{} with zero-shot prompting.
For closed questions, zero-shot prompting only achieves an average fidelity of 48\% with an average stability of 50\%.
Prompt engineering achieves an average fidelity of 97\% with an average stability of 85\%, because the underlying large language model has been provided with better information about the scope and nature of answers expected via initial prompts.

Indeed, prompt engineering together with low-temperature settings can lead to a fidelity of 100\% with a stability of 100\%, because the underlying large language model gives deterministic answers.
Here, increasing the temperature leads to a lower fidelity and lower stability, as the chance of wrong answers being created from an already strong baseline of correct explanations increases.

Interestingly, when considering zero-shot prompting, a temperature larger than $0$ leads to a higher fidelity albeit with the same low stability.
The higher temperature and the resulting more "creative" answers from the AI chatbot appear to increase the chance of providing a correct explanation as we start from a very weak baseline of correct explanations.

Finally, as one might expect, providing correct explanations for the open questions is more challenging.
Except for temperature = 0, the fidelity for open questions is generally lower than for closed questions (88\% vs. 97\% on average) and also exhibits a lower stability (74\% vs. 89\% on average), indicating a higher randomness of answers.

Comparing the fidelity of \app{} with the effectiveness of software engineers, \app{} was able to outperform the software engineers for 8 out of the 12 experiment configurations and was able to answer all eight questions correctly (100\% fidelity) in 3 configurations.
According to~\cite{TAAS2023}, only 33\% of the software engineers were able to answer all eight questions correctly.

\subsection{Validity Risks}
\label{sec:risks}
Concerning internal validity, we addressed the risk that results for \app{} may have been achieved by chance.
To this end, we carefully controlled experimental variables as introduced above.
In addition, we repeated the experiment multiple times to account for the typical stochastic effects and thus variance of machine learning models~\cite{PhamQWLRTYN20}.
While we chose metrics for \app{} and the user study that allow numerically comparing them, semantically they are not fully comparable.
To do so would require complementing the \app{} results by user studies that assess how well explainees could use the natural-language explanations to understand Deep RL decision making.

Concerning external validity, we chose a concrete service exemplar (SWIM) together with real-world workload traces and an actual subset of the interactions between Deep RL and the service environment.
Still, our experiments cover only one concrete problem in service-oriented systems (i.e., service adaptation) and use only one concrete service-oriented system exemplar.
We cover different styles of questions ("what/which", "why", "how many") reflecting various insights into the decision-making of Deep RL.
Yet, we limited the questions to the ones from the user study in~\cite{TAAS2023} to compare the performance of \app{} with that of software engineers.
While we designed \app{} to be as generic as possible and thus applicable to different problems in service-oriented systems, experimental results are limited with respect to generalizability.

%% file: discussion.tex
\section{Current Limitations and Potential Enhancements}
\label{sec:future}

\runin{Multi-round Question Answering}
Currently, \app{} generates a natural-language explanation for a given question.
Enhancing \app{} to also allow for follow-up questions thus appears as a natural next step.
An interesting further direction for such multi-round question answering is to follow the metaphor of the Socratic dialogue.
A Socratic dialogue may take the form of a cooperative argumentative dialogue between the explainer and explainee, where the explainer initiates the dialogue by asking questions to stimulate ideas by the explainee~\cite{WhiteEA2023}.
This may especially help non-technical users to come up with concrete questions concerning the decision-making of Deep RL.

\runin{Coping with Missing Insights}
Once an explainee is provided with the powerful natural-language interface of an AI chatbot, the explainee may ask questions concerning the decision-making of Deep RL for which the underlying explainable AI technique (here: XRL-DINE) does not provide the required insights.
One may check whether the answer given by the AI chatbot is backed by actual insights and inform the explainee accordingly. 
Here, work on explainable large language models may be leveraged\footnote{e.g., see \url{https://docs.aleph-alpha.com/docs/explainability/explainability/}}.


\runin{Protecting Sensitive Information}
AI chatbots and thus \app\ may be tricked into revealing information that should not be given away~\cite{HasalNSASO21,AIReport2023}.
One open question thus is how to leverage AI chatbots for explainable service-oriented systems while protecting sensitive information of the service provider~\cite{FolstadALBPRBLM21}.

\runin{Explanations for Policy-based Deep RL}
\app{} only works for value-based Deep RL because the underlying XRL-DINE technique needs access to the learned decision-making policy in the form of an action-value function $Q(S,A)$, giving the expected cumulative reward when executing action $A$ in state $S$.
In contrast to value-based Deep RL, policy-based Deep RL has the important advantage that it can naturally cope with concept drifts (e.g., see~\cite{CAISE2020,MoXZQL21,HuangWMCC22} for its use in service-oriented systems).
However, policy-based Deep RL does not use an action-value function $Q(S,A)$, because the fundamental idea is to directly use and optimize a parametrized stochastic action selection policy $\pi_\theta(S,A)$ in the form of a deep artificial neural network.

%% file: sota.tex
\section{Related Work}
\label{sec:sota}
As introduced in Sect.~\ref{sec:intro}, RL -- and recently Deep RL -- was successfully applied to different problems in service-oriented systems~\cite{RazianFBTB22,JamilISMB22,HuangWMCC22,YuZSLLWWL22,IftikharEa23,MutanuK19}.
While existing papers provide extensive and systematic evaluations of the performance of Deep RL for service-oriented systems, they did not yet address the problem of how to explain the decision-making of Deep RL using natural language.

There exists related work on natural-language explanations in the general area of explainable AI (XAI), which can be grouped in two main clusters.

\runin{Requirements Elicitation} Jentzsch et al. perform an empirical study to elicit user expectations towards chatbots for XAI~\cite{JentzschHH19}.
Results indicate that questions of users differ concerning the question form (open vs. closed), the level of abstraction, and the temporal scope (e.g., past, current or future outcomes).
Kuzba and Biecek elicit typical questions a human would ask a chatbot~\cite{KuzbaB20}.
From around 600 collected dialogues, they distill the 12 most common types of questions. 
Liao et al. construct a corpus of questions via literature review, as well as expert reviews and interviews~\cite{LiaoGM20}.
The resulting corpus contains 73 types of questions in 10 categories.
Gao et al. introduce a chatbot-based explanation framework built using IBM Watson Assistant~\cite{GaoLXA21}.
Similar to Kuzba and Biecek, they use this framework to understand what users would like to know about AI and elicit concrete user requests about AI-generated results.

All these works provide interesting insights into what questions may be asked, but are limited to supervised learning.
Also, in contrast to these works, \app{} does not require the up-front elicitation and fine-grained classification of questions.
\app{} works with two types of questions only because more specific  aspects are  handled naturally by the underlying powerful large language model.

\runin{Design and Realization} 
Carneiro et al. suggest combining a chatbot with an explainable model serving as a less complex surrogate than the actual black-box prediction model~\cite{Carneiro2021}.
They use natural language understanding to extract structured information from user questions concerning the user intent and the entity concerned with the question. 
Nguyen et al. leverage classical conversational agent architectures for question-answer dialogues~\cite{Nguyen2022}.
They (1) construct a question phrase bank, (2) establish a mapping between questions and explainable AI techniques, and (3) use template-based natural language generation to create explanations.
Malandri et al. consider the knowledge and experience of the users when generating explanations~\cite{MalandriMMN23}.
They explicitly introduce "clarification" as a further dialogue type on top of an earlier explanation framework~\cite{Madumal0SV19}.
Their user study indicated that different user groups perceive explanations differently, that all user groups prefer textual explanations over graphical ones, and that clarifications can enhance the usefulness of explanations.

All these works provide evidence for the benefits of using natural-language explanations.
However, they do not deliver explanations for Deep RL.
Also, they are all built using classical dialogue systems and natural language processing, and thus do not leverage the capabilities of modern large language models.

%% file: summary.tex
\section{Conclusion}
\label{sec:summary}

We took a first step towards natural-language explanations of Deep RL used for realizing service-oriented systems.
We introduced \app{}, an explainable AI technique powered by modern AI chatbot technology built on top of large language models.
We performed a proof-of-concept implementation for \app{} using ChatGPT.
Experimental results suggest that \app{} can provide explanations with high fidelity and stability.

In future work, we will work on the potential enhancements of \app{} described above.
We will also extend our experiments to other service-oriented system exemplars, which cover additional problems, such as service composition.
These experiments will be complemented by user studies to assess how useful and easy-to-use natural-language explanations are.

\runin{Acknowledgments}
We cordially thank Xhulja Shahini for her comments on earlier drafts of the paper.
Research leading to these results  received funding from the EU's Horizon Europe R\&I programme under grant 101070455 (DynaBIC).